\documentclass[lettersize,journal]{IEEEtran}

\usepackage{amsmath,amsfonts,amssymb}
\usepackage{algorithmic}
\usepackage{algorithm}
\usepackage{array}
\usepackage[caption=false,font=normalsize,labelfont=sf,textfont=sf]{subfig}
\usepackage{textcomp}
\usepackage{stfloats}
\usepackage{url}
\usepackage{verbatim}
\usepackage{graphicx}
\usepackage{cite}
\usepackage{svg}
\usepackage{multirow}
\usepackage{booktabs}
\usepackage{colortbl}
\usepackage{color}
\usepackage{xcolor}
\usepackage{soul}
\usepackage{bm}
\usepackage[breaklinks=true,colorlinks=true,linkcolor=black,citecolor=black,urlcolor=black]{hyperref}
\usepackage{cleveref}
\usepackage{siunitx}
\usepackage{balance}

\begin{document}

\title{SPEGNet: Synergistic Perception-Guided Network for Camouflaged Object Detection}

\author{Baber~Jan,~Saeed~Anwar,~Aiman~H.~El-Maleh,~Abdul~Jabbar~Siddiqui,~and~Abdul~Bais
\thanks{B. Jan, A. H. El-Maleh, and A. J. Siddiqui are with the Computer Engineering Department, King Fahd University of Petroleum and Minerals, Dhahran 31261, Saudi Arabia (e-mail: g202213840@kfupm.edu.sa; aimane@kfupm.edu.sa; abduljabbar.siddiqui@kfupm.edu.sa).}
\thanks{S. Anwar is with the Department of Computer Science and Software Engineering, The University of Western Australia, Perth, WA 6009, Australia (e-mail: saeed.anwar@uwa.edu.au).}
\thanks{A. Bais is with the Electronic Systems Engineering, University of Regina, Regina, SK S4S 0A2, Canada (e-mail: Abdul.Bais@uregina.ca).}
\thanks{Manuscript received Month XX, 2024; revised Month XX, 2024.}}


\markboth{IEEE Transactions on Image Processing,~Vol.~XX, No.~XX, Month~2024}%
{Jan \MakeLowercase{\textit{et al.}}: SPEGNet: Synergistic Perception-Guided Network for Camouflaged Object Detection}

\maketitle

\begin{abstract}
Camouflaged object detection segments objects with intrinsic similarity and edge disruption. Current detection methods rely on accumulated complex components. Each approach adds components such as boundary modules, attention mechanisms, and multi-scale processors independently. This accumulation creates a computational burden without proportional gains. To manage this complexity, they process at reduced resolutions, eliminating fine details essential for camouflage. We present SPEGNet, addressing fragmentation through a unified design. The architecture integrates multi-scale features via channel calibration and spatial enhancement. Boundaries emerge directly from context-rich representations, maintaining semantic-spatial alignment. Progressive refinement implements scale-adaptive edge modulation with peak influence at intermediate resolutions. This design strikes a balance between boundary precision and regional consistency. SPEGNet achieves 0.887 $S_\alpha$ on CAMO, 0.890 on COD10K, and 0.895 on NC4K, with real-time inference speed. Our approach excels across scales, from tiny, intricate objects to large, pattern-similar ones, while handling occlusion and ambiguous boundaries. Code, model weights, and results are available on \href{https://github.com/Baber-Jan/SPEGNet}{https://github.com/Baber-Jan/SPEGNet}.
\end{abstract}


\section{Introduction}
\label{sec:introduction}

\IEEEPARstart{C}{amouflaged} object detection (COD) identifies and segments objects that blend with their backgrounds through color, texture, and pattern similarity~\cite{fan2021concealed}. Unlike standard segmentation, COD targets objects specifically evolved or designed to minimize visual distinction. This similarity manifests through biological adaptation in nature and intentional concealment in artificial systems. The perceptual challenge requires specialized algorithms beyond general vision models. COD enables critical applications including medical polyp detection~\cite{fan2020pranet}, wildlife species monitoring~\cite{Wildlife}, and industrial defect identification~\cite{bhajantri2006camouflage}.

\begin{figure}[t]
    \centering
    \includegraphics[width=\columnwidth]{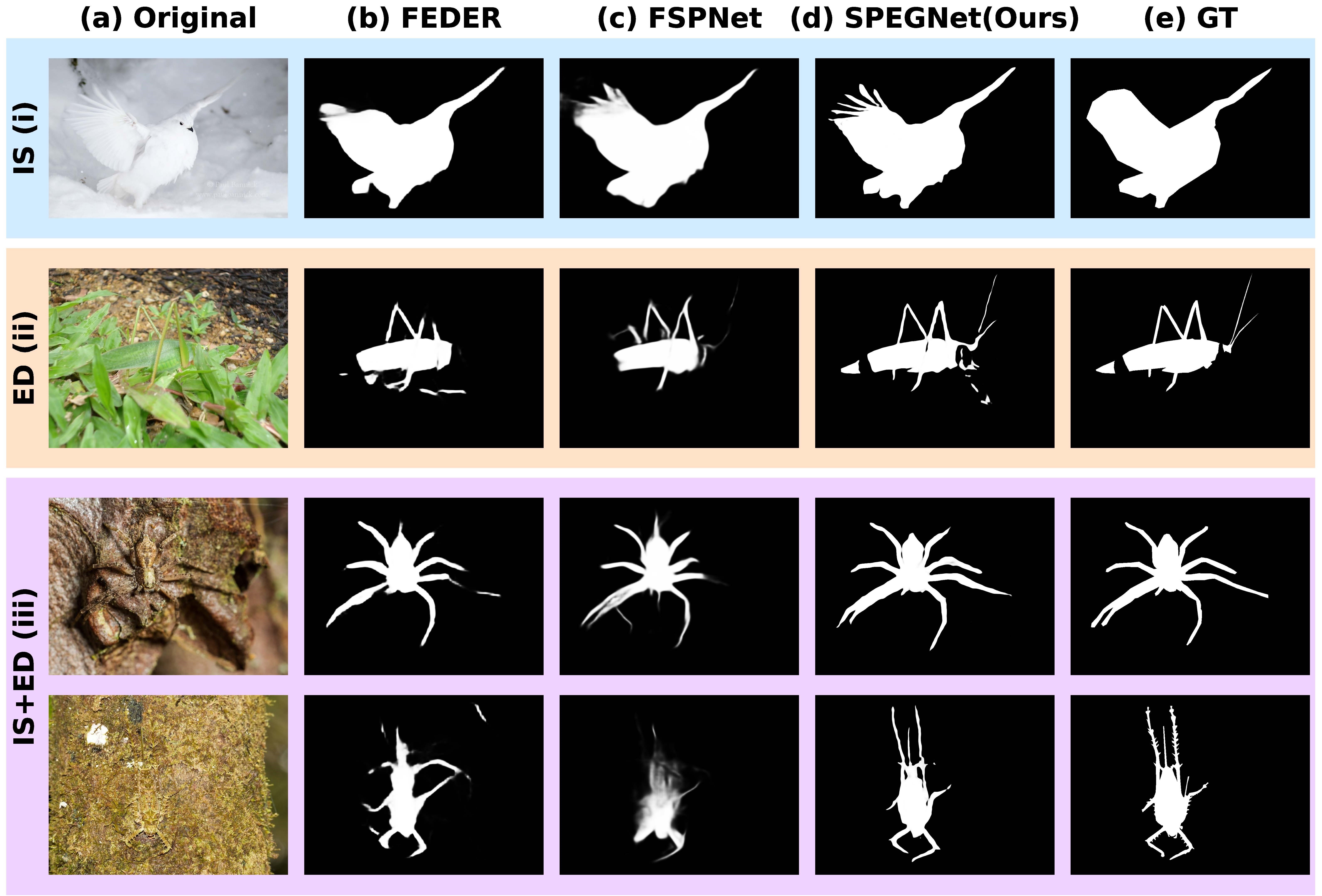}
    \caption{SPEGNet's effectiveness across diverse camouflage challenges. Columns show: (a) Original images, (b-d) Predictions from FEDER~\cite{he2023feder}, FSPNet~\cite{huang2023fspnet} and SPEGNet (Ours), (e) Ground truth. Rows show: (i) Intrinsic Similarity (IS)—white bird in snow, (ii) Edge Disruption (ED)—grasshopper with ambiguous boundaries, (iii-iv) Combined IS+ED with pattern similarity and intricate boundaries.}
    \label{fig:intro}
\end{figure}

Recent advances in general segmentation models~\cite{ravi2024sam2} fail to address camouflage-specific challenges~\cite{tang2024evaluatingsam2srolecamouflaged}. These models assume objects differ visually from backgrounds, but camouflage violates this assumption. COD faces two fundamental hurdles that explain this failure. Intrinsic Similarity (IS) occurs when objects match background appearance through evolved patterns. Edge Disruption (ED) fragments boundaries through texture continuity and gradual transitions. Fig.~\ref{fig:intro} demonstrates how IS and ED manifest across natural scenarios.

Current COD methods fragment detection through accumulated complex components, addressing COD challenges. CNN-based approaches~\cite{fan2020sinet,fan2021concealed,mei2021pfnet,Zhu_Li_Xie_Yan_Liang_Chen_Wei_Qin_2022,he2023feder} compensate limited receptive fields with feature pyramids, attention mechanisms, and boundary networks. In contrast, transformer methods~\cite{huang2023fspnet,pei2022osformer,yang2021ugtr} address tokenization detail loss with graph propagation, progressive refinement, and local pathways. Both paradigms inherit previous complexity while adding task-specific components. This architectural accumulation generates computational overhead without solving camouflage's fundamental coupling. Managing complexity requires processing at reduced resolutions, thereby eliminating texture gradients and boundary transitions that are essential for detecting IS and ED. Fig.~\ref{fig:intro}(b)-(c) demonstrate these systematic failures. Despite accumulated complexity, methods cannot reliably detect camouflaged objects.

We propose SPEGNet, built on synergistic perception principles for camouflaged object detection. Rather than accumulating modules to address failures, we design complementary components that work in concert with each other. Our CFI module combines channel recalibration with spatial context enhancement. These mechanisms jointly identify patterns that distinguish camouflaged objects from their backgrounds. The EFE module derives boundaries directly from enhanced contextual features. This preserves semantic-spatial correspondence, preventing the loss typically observed in edge networks. Our PED implements scale-adaptive edge modulation, concentrating refinement at intermediate resolutions. This targeted approach strikes a balance between precision and over-segmentation. Through synergistic design, SPEGNet avoids iterative accumulation, achieving state-of-the-art performance. Fig.~\ref{fig:intro}(d) shows that our results surpass those of complex accumulated architectures.

Our main contributions include:
\begin{itemize}
    \item A synergistic architecture achieving state-of-the-art COD performance without modular accumulation. SPEGNet reaches 0.890 $S_\alpha$ on COD10K, 0.895 on NC4K, and 0.887 on CAMO. It excels at intricate details, small and large pattern-similar objects, multiple instances, occlusions, and ambiguous boundaries.
    
    \item Channel recalibration combined with spatial pooling in a single module. This design amplifies object features while suppressing similar background patterns.
    
    \item Direct boundary extraction from contextual features, preserving semantic meaning. Edges know what object they belong to, preventing false boundaries in textured regions.
    
    \item Non-monotonic edge influence (20\%→33\%→0\%) across decoder stages. Peak influence at middle resolution captures camouflage boundaries most effectively.
\end{itemize}

\section{Related Work}
\label{sec:related_work}

Camouflaged object detection methods have evolved through the accumulation of architectural advancements. Each advancement adds computational components to address perceived limitations. This accumulation creates a processing burden without solving fundamental challenges. We review this progression from traditional approaches to current deep architectures.

\subsection{Evolution of Camouflaged Object Detection Methods}

Early COD methods used hand-crafted features with limited success. Bhajantri~\emph{et al.}~\cite{bhajantri2006camouflage} and Sengottuvelan~\emph{et al.}~\cite{sengottuvelan2008performance} analyzed texture through co-occurrence matrices. Likewise, only a limited number of methods~\cite{tankus2001convexity, pan2011study} detected convexity either from shading cues or from 3D structures. These techniques performed effectively on simple backgrounds but failed on complex camouflage patterns. This led to the development of deep learning methodologies that automatically learn features.

CNN-based methods established deep learning's entry into camouflaged object detection. Convolutional architectures excel at hierarchical feature learning but face fundamental limitations. Their local receptive fields struggle to capture large-scale camouflage patterns spanning entire image regions. Limited context understanding misses global relationships crucial for distinguishing camouflaged objects. To address these constraints, methodologies have progressively incorporated specialized components, such as the integration of search-identification networks~\cite{fan2020sinet}, the utilization of positioning modules~\cite{mei2021pfnet} for small objects detection, the application of cross-level fusion~\cite{sun2021c2fnet} for multi-scale processing, the employment of bilateral attention~\cite{Zhu_Li_Xie_Yan_Liang_Chen_Wei_Qin_2022} for boundary-region modeling, the stacking of boundary networks~\cite{sun2022bgnet} for edge enhancement, and the development of graph modules~\cite{zhai2021Mutual} for relationship modeling.
Despite these additions, performance plateaus across methods. The accumulated components operate within the local processing constraints of the CNN. Each module addresses symptoms rather than the core limitation. Camouflage detection requires a unified understanding of both global and local contexts, which fragmented CNN architectures fail to provide.

Transformer architectures were introduced into COD to address the limitations of CNNs regarding their local receptive fields. Self-attention mechanisms are highly effective in modeling global dependencies across entire images. Nonetheless, transformers encounter fundamental challenges when applied to camouflaged object detection. Token-based processing tends to lose essential spatial structures that are intrinsic to camouflage patterns. Global attention mechanisms blur local boundary details essential for edge-disrupted objects. Quadratic complexity restricts practical processing to lower resolutions. To overcome these challenges, researchers have progressively used hybrid components, e.g., appending CNN backbones~\cite{yang2021ugtr} to preserve local features, introducing one-shot modules~\cite{pei2022osformer} to maintain spatial resolution, incorporating feature shrinkage paths~\cite{huang2023fspnet} with graph convolutions, and integrating hierarchical features~\cite{hu2023high} between CNN and transformer models. 
Each addition attempts to compensate for transformer limitations. Yet these hybrid architectures multiply complexity without solving core issues. The accumulated components struggle to reconcile global attention with spatial precision. Methods achieve similar performance despite vastly different designs. Transformers excel at global modeling but struggle to provide a unified spatial-semantic understanding for camouflage.

\subsection{Multi-Scale Processing Architectures}  

Multi-scale processing emerged to address scale variation in camouflaged objects. Small creatures hide through fine texture mimicry while large animals blend via global pattern matching. Single-scale features fail to capture critical detection cues at varying sizes. Fixed receptive fields cannot simultaneously capture local details and global context. To address these constraints, methods have progressively accumulated parallel processing pathways. For example, FPN~\cite{lin2017feature} establishes top-down connections that combine coarse semantics with fine spatial features, and PANet~\cite{liu2018path} introduces bottom-up pathways to enable bidirectional information flow. Meanwhile, C2FNet~\cite{sun2021c2fnet} develops dense cross-level connections among all scale pairs, and ZoomNet~\cite{pang2022zoom} uses progressive scale expansion with specialized modules. Lastly, Huang~\emph{et al.}~\cite{huang2023fspnet} creates hierarchical shrinkage paths that connect adjacent scales. Each addition increases the number of architectural pathways and memory usage. Despite the proliferation of pathways, performance improvements remain marginal. The accumulated pathways process scales independently before attempting fusion. This fragmentation cannot model camouflage patterns that span multiple scales simultaneously. Multi-scale accumulation perpetuates the same fundamental limitation.

\subsection{Edge Detection and Boundary Refinement}

Edge disruption presents a fundamental challenge in camouflaged object detection. Camouflage deliberately fragments boundaries through texture continuity and gradual color transitions. Traditional edge detection methods fail when boundaries blend seamlessly with their backgrounds. Clear object delineation requires understanding both semantic content and spatial boundaries simultaneously. To address boundary ambiguity, methods progressively accumulated specialized edge processing components. For example, EGNet~\cite{zhao2019egnet} created dual-stream architectures that separate edge and region processing, and BSANet~\cite{Zhu_Li_Xie_Yan_Liang_Chen_Wei_Qin_2022} incorporated boundary-sensitive attention modules to enhance contours. According to He~\emph{et al.}~\cite{he2023feder}, appending ODE-inspired refinement models to boundary evolution across scales and Liu~\emph{et al.}~\cite{liu2022boosting} added transformer-based interactive branches for boundary-semantic fusion. Similarly, BGNet~\cite{sun2022bgnet} stacked explicit boundary guidance networks atop segmentation streams. Each method fragments detection into parallel pathways that process edges independently. The separated streams lose crucial semantic-edge relationships. Edge extraction without object understanding generates false responses in textured regions. Semantic processing without edge awareness produces imprecise boundaries. Late-stage fusion cannot reconcile this fundamental disconnection. The accumulated edge components perpetuate fragmentation rather than achieving a unified understanding of the boundary.


The evolution of camouflaged object detection reveals a consistent pattern across three generations. Traditional methods accumulated hand-crafted features targeting isolated aspects. CNN architectures have accumulated specialized modules that address local processing constraints. Transformer approaches integrate hybrid components, striking a balance between global and local understanding. Multi-scale processing enables the accumulation of parallel pathways for comprehensive scale coverage. Edge methods accumulated dual streams for boundary refinement. Each generation inherited previous complexity while adding new components. This architectural accumulation creates fragmented processing, unable to model camouflage's unified nature. Intrinsic similarity and edge disruption occur simultaneously, requiring integrated perception. Yet, accumulated architectures process these challenges through independent pathways. Performance plateaus across vastly different designs confirm this fundamental limitation. The field requires a paradigm shift from accumulation to synergy. Next, we present our SPEGNet's alternative approach through complementary component design.
\section{Methodology}
\label{sec:methodology}

\begin{figure*}[tbp!]
    \centering
    \includegraphics[width=\textwidth]{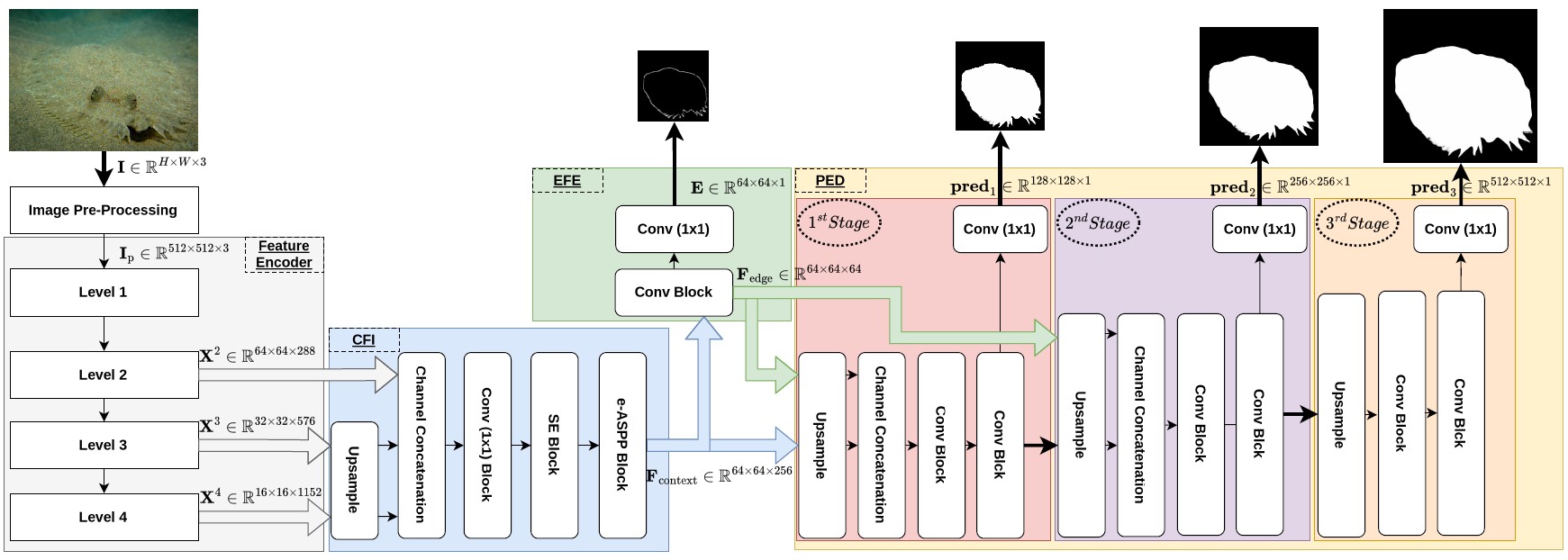}
    \caption{Architecture overview of SPEGNet. The figure illustrates the data flow through four key components: Feature Encoding (gray), Contextual Feature Integration (blue), combining multi-scale features, Edge Feature Extraction (green), deriving boundary information, and Progressive Edge-guided Decoder (yellow), which generates multi-scale predictions with scale-adaptive edge modulation. Sample input and corresponding segmentation outputs at different refinement stages are shown at the top of the figure.}
    \label{fig:architecture}
\end{figure*}

Given an input image $\mathbf{I} \in \mathbb{R}^{H \times W \times 3}$ containing camouflaged objects, our goal is to generate a binary segmentation mask $\mathbf{M} \in \{0,1\}^{H \times W}$. The mask must precisely delineate object boundaries despite intrinsic similarity and edge disruption. Input image $\mathbf{I}$ undergoes preprocessing (resizing and normalization) to produce $\mathbf{I}_p \in \mathbb{R}^{H \times W \times 3}$. Throughout this work, we use the following notation: $\mathbf{X}^s$ for features at encoder stage $s \in \{1,2,3,4\}$, $\mathbf{F}_\text{context}$ for context-enhanced features, $\mathbf{F}_\text{edge}$ for edge features, $\mathbf{E}$ for edge predictions, $\mathbf{P}_i$ for decoder predictions at stage $i$, and $\mathbf{D}_i$ for decoder features. Our SPEGNet processes camouflaged images through synergistic modules. A hierarchical vision transformer encoder transforms $\mathbf{I}_p$ into multi-scale features $\{\mathbf{X}^1, \mathbf{X}^2, \mathbf{X}^3, \mathbf{X}^4\}$. Our network introduces three complementary modules to process these features: (i) Contextual Feature Integration, which transforms multi-scale features into discriminative features; (ii) Edge Feature Extraction, which derives boundary information and edge predictions; and (iii) Progressive Edge-guided Decoder, which combines all features to produce the final mask, as shown in Fig.~\ref{fig:architecture}. More details are provided in the upcoming sections. 


\subsection{Feature Encoding.}
\label{subsec:fe}
We employ Hiera-Large~\cite{ryali2023hiera} as our vision transformer encoder. The encoder transforms preprocessed image $\mathbf{I}_p$ into multi-scale features $\{\mathbf{X}^s\}_{s=1}^4$. Each stage $s$ produces $\mathbf{X}^s \in \mathbb{R}^{H_s \times W_s \times C_s}$ where $[H_s, W_s] = [H/2^{s+1}, W/2^{s+1}]$ and $C_s \in \{144, 288, 576, 1152\}$. Spatial dimensions halve while channels double at each stage. We utilize features from stages 2-4 for subsequent processing. Stage 2 preserves boundary details, stage 3 balances spatial and semantic information, and stage 4 captures high-level understanding. These multi-scale features are then fed into our three complementary modules.

\subsection{Contextual Feature Integration (CFI)}
\label{subsec:cfi}

Camouflaged objects exhibit intrinsic similarity through two mechanisms: channel ambiguity, where certain features match the background, and spatial confusion, where patterns repeat across different scales. Addressing both challenges requires unified processing that enhances discriminative features while suppressing ambiguous ones. CFI transforms encoder features $\{\mathbf{X}^2, \mathbf{X}^3, \mathbf{X}^4\}$ into context-enhanced features $\mathbf{F}_{\text{context}} \in \mathbb{R}^{H/8 \times W/8 \times 256}$ through integrated channel-spatial processing. The module first aligns features by upsampling stages 3 and 4 to the resolution of stage 2, then concatenates them. After concatenation, the module performs channel recalibration using the squeeze-and-excitation method~\cite {hu2018squeeze} to identify discriminative features. After channel recalibration, the module performs spatial enhancement using e-ASPP~\cite{MODNet} to capture multi-scale patterns producing $\mathbf{F}_{\text{context}}$. This synergistic approach produces discriminative representations by simultaneously addressing both intrinsic similarity aspects, enabling effective camouflage detection in subsequent modules.

\subsection{Edge Feature Extraction (EFE)}
\label{subsec:efe}

Edge disruption in camouflage creates fragmented boundaries where objects blend seamlessly with backgrounds. Detecting these subtle boundaries requires maintaining semantic understanding throughout edge extraction. EFE receives context-enhanced features $\mathbf{F}_{\text{context}}$ from CFI and transforms them into edge features $\mathbf{F}_{\text{edge}} \in \mathbb{R}^{H/8 \times W/8 \times 64}$ and edge predictions $\mathbf{E} \in \mathbb{R}^{H/8 \times W/8 \times 1}$. EFE begins by encoding context features using convolutional layers to extract boundary-aware representations $\mathbf{F}_{\text{edge}}$. The module then derives edge predictions $\mathbf{E}$ from these edge features through a 1$\times$1 convolution. Edge features $\mathbf{F}_{\text{edge}}$ guide boundary refinement in the decoder while predictions $\mathbf{E}$ provide edge supervision. This semantic-preserving extraction ensures boundaries align with actual objects rather than texture patterns.

\subsection{Progressive Edge-guided Decoder (PED)}
\label{subsec:ped}

Multi-scale refinement requires balancing boundary precision with region consistency across different resolutions. Coarse features capture the presence of objects, while fine features reveal boundary details. PED receives context features $\mathbf{F}_{\text{context}}$ from CFI and edge features $\mathbf{F}_{\text{edge}}$ from EFE. The decoder transforms these features into the final mask $\mathbf{M}$ through a three-stage progressive refinement process. The decoder first produces prediction $\mathbf{P}_1$ at quarter resolution with 20\% edge influence to establish initial object localization. After initial refinement, the decoder generates $\mathbf{P}_2$ at half resolution with 33\% edge influence to maximize boundary precision. After intermediate refinement, the decoder produces $\mathbf{P}_3$ at full resolution with 0\% edge influence to ensure region consistency. This peak-and-fade modulation concentrates edge guidance at intermediate scales, where it is most effective. The Bayesian-inspired refinement treats coarse predictions as priors updated by finer evidence. The final mask $\mathbf{M}$ accurately delineates camouflaged objects through this progressive approach.

\subsection{Loss Functions}
\label{subsec:loss}

Training SPEGNet requires supervising multiple outputs across different scales and tasks. Each component produces predictions requiring appropriate supervision strategies. Our loss function combines segmentation and edge objectives through weighted summation:
\begin{equation}
    \mathcal{L}_{\text{total}} = \sum_{i=1}^{3} w_i \mathcal{L}_{\text{seg}}(\mathbf{P}_i, \mathbf{M}_{\text{gt}}) + \lambda_e \mathcal{L}_{\text{edge}}(\mathbf{E}, \mathbf{E}_{\text{gt}})
\end{equation}
where $w_i$ implements progressive weighting across decoder stages and $\lambda_e$ balances edge supervision. The segmentation loss $\mathcal{L}_{\text{seg}}$ combines boundary-aware BCE and IoU terms. BCE performs pixel-wise binary classification distinguishing object from background. IoU optimizes region-level overlap to address intrinsic similarity challenges. Boundary-aware weighting emphasizes transitional regions where camouflage creates ambiguity. The edge loss $\mathcal{L}_{\text{edge}}$ employs focal and dice components. Focal loss handles severe foreground-background imbalance in edge maps. Dice loss ensures continuous boundary formation despite edge disruption. Progressive weighting concentrates supervision on finer resolutions where details matter most. This multi-objective formulation ensures balanced learning across all scales and tasks. The combined supervision enables SPEGNet to learn both accurate boundaries and consistent regions simultaneously.
\section{Experiments}
\label{sec:experiments}

We evaluate SPEGNet on three benchmark datasets and analyze its performance across multiple dimensions. This section presents implementation details, quantitative comparisons, resolution analysis, ablation studies, and domain transfer capabilities.

\subsection{Experimental Setup}

\paragraph{Implementation Details.} Input images are preprocessed by resizing to 512$\times$512 using bilinear interpolation, normalizing to [0,1], and applying ImageNet normalization ($\boldsymbol{\mu}$=[0.485, 0.456, 0.406], $\boldsymbol{\sigma}$=[0.229, 0.224, 0.225]). The implementation utilizes the PyTorch framework on an NVIDIA H100 GPU with 93.6GB of memory and a Hiera-Large backbone from SAM2~\cite{ravi2024sam2}. Training runs for 150 epochs using the AdamW optimizer (learning rate 1e$^{-4}$, weight decay 1e$^{-5}$) with a batch size of 42. Pre-trained encoder layers use a learning rate of 5e$^{-5}$. The ReduceLROnPlateau scheduler uses a factor of 0.7, patience of 5, and a minimum of 1e$^{-6}$. Training employs mixed-precision computation and gradient clipping with a threshold of 1.0. Progressive scale weights are [0.2, 0.3, 0.5]. Structure loss weights are $\lambda_{\text{bce}}$=1.25, $\lambda_{\text{iou}}$=1.0, and $\lambda_b$=2.0. Edge loss parameters are $\lambda_e$=0.75, $\alpha$=0.75, and $\gamma$=2.0. Training reserves 10\% data for validation. Evaluation processes single images, maintaining identical preprocessing.

\paragraph{Datasets.} We evaluate on three established COD benchmarks widely adopted in the literature. All datasets provide pixel-level ground truth annotations. Edge maps are generated using the Canny detector on ground truth masks with a 5-pixel width for training supervision.
\begin{itemize}
    \item \textbf{COD10K}~\cite{fan2021concealed}: Largest COD dataset containing 10,000 images with 5,066 camouflaged instances. The training set has 3,040 images, and the testing set comprises 2,026 images. The dataset encompasses 10 superclasses and 78 subclasses, spanning aquatic, terrestrial, flying, and amphibian categories.
    
    \item \textbf{CAMO}~\cite{le2019anabranch}: Contains 1,250 images featuring both natural and artificial camouflage patterns. Training utilizes 1,000 images, while testing employs 250 images. The dataset includes eight categories with diverse camouflage strategies, including color matching, pattern mimicry, and background blending.
    
    \item \textbf{NC4K}~\cite{NC4K}: Test-only benchmark containing 4,121 images with challenging natural camouflage scenarios. This dataset evaluates cross-dataset generalization without fine-tuning. Images feature extreme background similarity testing detection limits.
\end{itemize}

\paragraph{Evaluation Metrics.} We employ five standard COD metrics widely adopted in the literature for comprehensive evaluation. All metrics except MAE use the higher is better convention.
\begin{itemize}
    \item \textbf{Structure Measure ($S_{\alpha}$)}~\cite{Cheng2021sMeasure}: Evaluates both region-aware and object-aware structural similarity. The metric captures structural integrity beyond pixel-level accuracy with $\alpha$=0.5 balancing object and region scores.
    
    \item \textbf{Enhanced Alignment Measure ($E_{\phi}$)}~\cite{fan2018enhanced}: Combines local pixel-level matching with global image statistics. The enhanced formulation addresses bias and sensitivity issues in the original version through improved statistical alignment.
    
    \item \textbf{Weighted F-Measure ($F_{\beta}^w$)}~\cite{margolin2014evaluate}: Assigns importance-based weights to different image regions, emphasizing errors in salient areas. Following COD evaluation protocols, $\beta^2$=0.3 balances precision and recall.
    
    \item \textbf{Mean F-Measure ($F_{\beta}^m$)}~\cite{meanFMeasure}: Averages F-measure across all thresholds from 0 to 255, providing threshold-independent assessment. This complements adaptive metrics with a comprehensive evaluation.
    
    \item \textbf{Mean Absolute Error ($\mathcal{M}$)}: Computes average pixel-wise absolute difference between prediction and ground truth. Lower values indicate better performance with direct error interpretation.
\end{itemize}

\subsection{Comparison with State-of-the-Art}

\paragraph{Quantitative Results.} Table~\ref{tab:comparison} evaluates SPEGNet against 27 state-of-the-art methods spanning CNN-based approaches (highlighted in gray) and transformer-based architectures. The comparison includes early CNN methods (SINet, PFNet), recent boundary-focused approaches (BGNet, FEDER), and transformer methods (FSPNet, HitNet). All results use author-provided predictions, ensuring fair comparison. On COD10K, SPEGNet achieves state-of-the-art 0.890 $S_{\alpha}$, surpassing FocusDiffuser (0.875) by 1.7\% and FSEL (0.873) by 2.0\%. Furthermore, the $F_{\beta}^w$ improvement reaches 3.7\% (0.839 vs 0.809), validating superior boundary preservation. Additionally, SPEGNet achieves the highest $E_{\phi}$ (0.949), demonstrating enhanced structural alignment. Similarly, NC4K shows strong cross-dataset generalization with 0.895 $S_{\alpha}$ and 0.860 $F_{\beta}^w$, outperforming all methods. Moreover, the $E_{\phi}$ score of 0.947 surpasses the second-best 0.941 by 0.6\%, confirming robust feature learning. Meanwhile, CAMO achieves the highest absolute performance with 0.887 $S_{\alpha}$, marginally surpassing FSEL (0.885) and FocusDiffuser (0.881). The $F_{\beta}^w$ improvement is more pronounced at 0.870 versus 0.851, demonstrating a 2.2\% gain. However, modest improvements on CAMO partially reflect annotation quality, where ground truth captures only single instances despite multiple objects present. Finally, general segmentation models demonstrate fundamental COD limitations with SAM-Auto achieving only 0.684 $S_{\alpha}$ and SAM2-Auto dropping to 0.444. These results confirm that specialized architectures outperform general segmentation by significant margins.

\begin{table*}[t]
   \centering
   \caption{Quantitative comparison with 27 SOTA methods on benchmark datasets. 
   Notes: $\uparrow$/$\downarrow$ denotes the larger/smaller value is better. 
   "-" indicates unavailable data. 
   The best values are in \textbf{\color{red}bold red}, the second best are \underline{\color{blue}underlined blue}. 
   Light gray rows are CNN-based methods; white rows are Transformer-based methods. All comparison results are obtained from author-provided predictions for fair comparison.}
   \label{tab:comparison}
   \resizebox{\textwidth}{!}{
   \begin{tabular}{l|ccccc|ccccc|ccccc}
       \toprule
       \multirow{2}{*}{Methods} 
       & \multicolumn{5}{c|}{CAMO (250)} 
       & \multicolumn{5}{c|}{COD10K (2,026)} 
       & \multicolumn{5}{c}{NC4K (4,121)} \\
       \cline{2-16}
       & $S_{\alpha}\uparrow$ & $F_\beta^w\uparrow$ & $F_\beta^m\uparrow$ & $E_\phi\uparrow$ & $\mathcal{M}\downarrow$
       & $S_{\alpha}\uparrow$ & $F_\beta^w\uparrow$ & $F_\beta^m\uparrow$ & $E_\phi\uparrow$ & $\mathcal{M}\downarrow$
       & $S_{\alpha}\uparrow$ & $F_\beta^w\uparrow$ & $F_\beta^m\uparrow$ & $E_\phi\uparrow$ & $\mathcal{M}\downarrow$ \\
       \midrule
       \rowcolor{gray!15}
       SINet$_{20}$~\cite{fan2020sinet} 
         & .751 & .606 & .675 & .831 & .100 
         & .771 & .551 & .634 & .868 & .051
         & .808 & .723 & .769 & .883 & .058 \\
       \rowcolor{gray!15}
       SLSR$_{21}$~\cite{NC4K}
         & .787 & .674 & .744 & .854 & .080
         & .804 & .673 & .715 & .892 & .037
         & .840 & .766 & .804 & .907 & .048 \\
       \rowcolor{gray!15}
       PFNet$_{21}$~\cite{mei2021pfnet}
         & .782 & .695 & .746 & .855 & .085
         & .800 & .660 & .701 & .890 & .040
         & .829 & .745 & .784 & .898 & .053 \\
       \rowcolor{gray!15}
       MGL$_{21}$~\cite{zhai2021Mutual}
         & .775 & .673 & .726 & .842 & .088
         & .814 & .666 & .711 & .890 & .035
         & .833 & .740 & .782 & .893 & .052 \\
       \rowcolor{gray!15}
       UJSC$_{21}$~\cite{li2021uncertainty}
         & .800 & .728 & .772 & .873 & .073
         & .809 & .684 & .721 & .891 & .035
         & .842 & .771 & .806 & .907 & .047 \\
       \rowcolor{gray!15}
       C$^2$FNet$_{21}$~\cite{sun2021c2fnet}
         & .796 & .719 & .762 & .864 & .080
         & .813 & .686 & .723 & .900 & .036
         & .838 & .762 & .795 & .904 & .049 \\
       \rowcolor{gray!15}
       UGTR$_{21}$~\cite{yang2021ugtr}
         & .784 & .684 & .735 & .851 & .086
         & .817 & .666 & .712 & .890 & .036
         & .839 & .747 & .787 & .899 & .052 \\
       \rowcolor{gray!15}
       PreyNet$_{22}$~\cite{10.1145/3503161.3548178}
         & .790 & .708 & .757 & .857 & .077
         & .813 & .697 & .736 & .891 & .034
         & - & - & - & - & - \\
       \rowcolor{gray!15}
       BSA-Net$_{22}$~\cite{Zhu_Li_Xie_Yan_Liang_Chen_Wei_Qin_2022}
         & .794 & .717 & .763 & .867 & .079
         & .818 & .699 & .738 & .901 & .034
         & .841 & .771 & .808 & .907 & .048 \\
       \rowcolor{gray!15}
       OCE-Net$_{22}$~\cite{9706783}
         & .802 & .723 & .766 & .865 & .080
         & .827 & .707 & .741 & .905 & .033
         & .853 & .785 & .818 & .913 & .045 \\
       \rowcolor{gray!15}
       BGNet$_{22}$~\cite{sun2022bgnet}
         & .812 & .749 & .789 & .882 & .073
         & .831 & .722 & .753 & .911 & .033
         & .851 & .788 & .820 & .916 & .044 \\
       \rowcolor{gray!15}
       SegMaR$_{22}$~\cite{Jia_2022_CVPR}
         & .815 & .795 & .794 & .884 & .071
         & .833 & .724 & .757 & .906 & .034
         & .841 & .781 & .820 & .907 & .046 \\
       \rowcolor{gray!15}
       ZoomNet$_{22}$~\cite{pang2022zoom}
         & .820 & .752 & .794 & .892 & .066
         & .830 & .729 & .766 & .911 & .029
         & .853 & .784 & .818 & .912 & .043 \\
       \rowcolor{gray!15}
       SINet-v2$_{22}$~\cite{fan2021concealed}
         & .820 & .743 & .782 & .895 & .070
         & .815 & .680 & .718 & .906 & .037
         & .847 & .770 & .805 & .914 & .048 \\
       \rowcolor{gray!15}
       FDNet$_{22}$~\cite{zhong2022detecting}
         & .828 & .748 & .781 & .883 & .068
         & .832 & .706 & .733 & .907 & .033
         & .834 & .750 & .784 & .893 & .051 \\
       \rowcolor{gray!15}
       DTINet$_{22}$~\cite{liu2022boosting}
         & .856 & .796 & - & .916 & .050
         & .824 & .695 & - & .896 & .034
         & .863 & .792 & - & .917 & .041 \\
       OSFormer$_{22}$~\cite{pei2022osformer}
         & .799 & - & - & .858 & .073
         & .811 & - & - & .881 & .034
         & .832 & - & - & .905 & .049 \\
       FSPNet$_{23}$~\cite{huang2023fspnet}
         & .856 & .799 & .830 & .928 & .050
         & .851 & .735 & .769 & .930 & .026
         & .879 & .816 & .843 & .937 & .035 \\
       TPRNet$_{22}$~\cite{zhang2023tprnet}
         & .814 & .781 & - & - & .076
         & .829 & .725 & - & - & .034
         & .854 & .790 & - & - & .047 \\
       FPNet$_{23}$~\cite{cong2023frequency}
         & .852 & .806 & - & .905 & .056
         & .850 & .748 & - & .913 & .029
         & - & - & - & - & - \\
       OPNet$_{23}$~\cite{mei2023camouflaged}
         & .858 & .817 & - & .915 & .050
         & .857 & .767 & - & .919 & .026
         & .883 & .838 & - & .932 & .034 \\
       HitNet$_{23}$~\cite{hu2023high}
         & .844 & .801 & - & .902 & .057
         & .868 & .798 & - & .932 & .024
         & .870 & .825 & - & .921 & .039 \\
       SAM-Auto$_{23}$~\cite{sam2023,tang2023samsegmentanythingsam}
         & .684 & .606 & .680 & .687 & .132
         & .783 & .701 & .756 & .798 & .050
         & .767 & .696 & .752 & .776 & .078 \\
       SAM-Prompt$_{23}$~\cite{sam2023,tang2023samsegmentanythingsam}
         & .647 & .520 & - & - & .141
         & .696 & .552 & - & - & .094
         & .699 & .591 & - & - & .115 \\
       \rowcolor{gray!15}
       FEDER$_{23}$~\cite{he2023feder}
         & .807 & .785 & \underline{\color{blue}{.873}} 
         & \textbf{\color{red}{.947}} 
         & .069
         & .823 & .740 & \textbf{\color{red}{.900}} & .911 & .032
         & .846 & .817 & \textbf{\color{red}{.905}} & .916 & .045 \\
       SAM2-Auto$_{24}$~\cite{ravi2024sam2,tang2024evaluatingsam2srolecamouflaged}
         & .444 & .184 & .207 & .401 & .236
         & .549 & .271 & .291 & .521 & .134
         & .512 & .251 & .268 & .482 & .186 \\
       SAM2-Prompt$_{24}$~\cite{ravi2024sam2,tang2024evaluatingsam2srolecamouflaged}
         & .722 & .633 & - & - & .114
         & .754 & .640 & - & - & .078
         & .776 & .700 & - & - & .085 \\
       FocusDiffuser$_{25}$~\cite{zhao2025focus}
         & .881 & \underline{\color{blue}{.851}} & - & .939 & .042
         & \underline{\color{blue}{.875}} & \underline{\color{blue}{.809}} & - & \underline{\color{blue}{.939}} & \textbf{\color{red}{.020}}
         & .891 & \underline{\color{blue}{.854}} & - & .940 & \underline{\color{blue}{.029}} \\
       FSEL$_{25}$~\cite{sun2025fsel}
         & \underline{\color{blue}{.885}} & \underline{\color{blue}{.851}} & .864 & .942 & \underline{\color{blue}{.040}}
         & .873 & .800 & .796 & .928 & \underline{\color{blue}{.021}}
         & \underline{\color{blue}{.892}} & .853 & .864 & \underline{\color{blue}{.941}} & .030 \\
       \hline
       \textbf{Ours}
         & \textbf{\color{red}{.887}}
         & \textbf{\color{red}{.870}}
         & \textbf{\color{red}{.882}}
         & \underline{\color{blue}{.943}}
         & \textbf{\color{red}{.037}}
         & \textbf{\color{red}{.890}}
         & \textbf{\color{red}{.839}}
         & \underline{\color{blue}{.847}}
         & \textbf{\color{red}{.949}}
         & \textbf{\color{red}{.020}}
         & \textbf{\color{red}{.895}}
         & \textbf{\color{red}{.860}}
         & \underline{\color{blue}{.870}}
         & \textbf{\color{red}{.947}}
         & \textbf{\color{red}{.025}} \\
       \bottomrule
   \end{tabular}
   }
   \vspace{-3mm}
\end{table*}

\begin{figure*}[t]
    \centering
    \includegraphics[width=\textwidth]{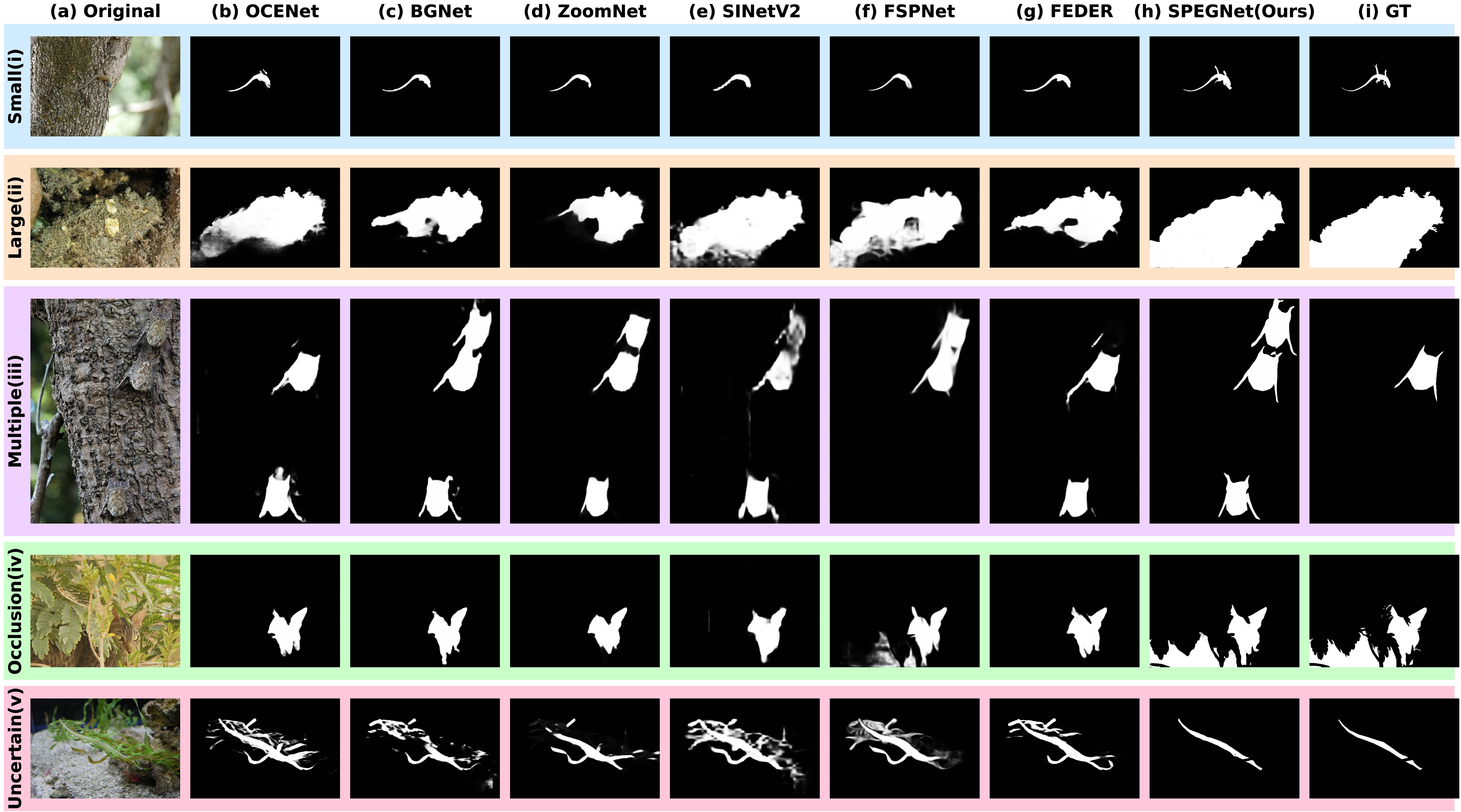}
    \caption{Qualitative comparison on challenging COD scenarios. Columns show: (a) input image, (b) ground truth, (c-h) predictions from OCENet~\cite{9706783}, BGNet~\cite{sun2022bgnet}, ZoomNet~\cite{pang2022zoom}, SINetV2~\cite{fan2021concealed}, FSPNet~\cite{huang2023fspnet}, FEDER~\cite{he2023feder}, and (i) SPEGNet (Ours). Rows demonstrate: (i) small object detection, (ii) large object with pattern similarity, (iii) multiple instances revealing annotation limitations, (iv) occlusion handling, and (v) ambiguous boundaries. SPEGNet consistently outperforms existing methods across all scenarios.}
    \label{fig:visual_comparison}
\end{figure*}

\paragraph{Qualitative Analysis.} Fig.~\ref{fig:visual_comparison} presents visual comparisons across five challenging scenarios comparing SPEGNet with OCENet, BGNet, ZoomNet, SINetV2, FSPNet, and FEDER. First, small, camouflaged objects (Fig.~\ref{fig:visual_comparison} row i) pose a challenge for most methods, resulting in fragmented or missed detections. Specifically, other methods produce incomplete boundaries and missing critical details. In contrast, SPEGNet precisely segments the complete insect, preserving fine structures through multi-scale context enhancement. Second, large objects with background similarity (Fig.~\ref{fig:visual_comparison} row ii) cause severe undersegmentation across competing methods. For instance, ZoomNet misses around 60\% of the fish body. However, SPEGNet generates a complete mask with clear boundaries as CFI's channel attention suppresses water patterns while enhancing fish-specific features. Third, multiple object scenarios (Fig.~\ref{fig:visual_comparison} row iii) reveal critical annotation limitations where ground truth marks only one instance despite three visible instances. Consequently, most methods detect all three instances but with incomplete boundaries. 

Interestingly, FSPNet, OCENet, and FEDER identify two birds that demonstrate limited multi-instance capability and some form of learning the labels, rather than generalization. Remarkably, SPEGNet correctly identifies all three birds with precise individual boundaries, demonstrating superior detection beyond training supervision. This observation highlights how CAMO's single-instance annotations significantly underestimate actual performance improvements. Fourth, occlusion handling (Fig.~\ref{fig:visual_comparison}, row iv) reveals significant differences in shape completion abilities. Almost all the methods only capture the head part of the animal and miss the lower body entirely. Conversely, SPEGNet reconstructs the complete shape, maintaining smooth contours through progressive refinement, leveraging contextual understanding. Finally, ambiguous boundary cases (Fig.~\ref{fig:visual_comparison}, row v) demonstrate variations in precision across methods. Here, all competing methods produce irregular, fragmented boundaries with significant oversegmentation or undersegmentation. In particular, these methods fail to accurately capture the precise shape of the stick insect against textured backgrounds. Nevertheless, our scale-adaptive edge modulation delineates exact boundaries, distinguishing subtle texture transitions. Ultimately, the peak influence at intermediate scales balances boundary precision with region consistency, achieving accurate segmentation.

\subsection{Resolution Impact Study}

\paragraph{Quantitative Analysis.} Resolution critically impacts camouflage detection as subtle discriminative features disappear during downsampling. Table~\ref{tab:resolution} evaluates SPEGNet across three resolutions (384$\times$384, 512$\times$512, 1024$\times$1024). Performance consistently improves from 384 to 512 resolution across all datasets. Furthermore, the boundary-sensitive $F_\beta^w$ metric shows notable gains (4.7\% on COD10K). Additionally, 1024 resolution yields substantial improvements on larger datasets. Specifically, COD10K achieves 0.908 $S_\alpha$ (2\% gain) and 0.867 $F_\beta^w$ (3.3\% gain). Similarly, NC4K reaches 0.903 $S_\alpha$ and 0.877 $F_\beta^w$. However, CAMO shows a slight decrease at 1024 resolution (0.884 vs 0.887 $S_\alpha$). This pattern reflects increased detection of non-annotated objects.

\begin{table}[t]
    \centering
    \caption{Performance at different input resolutions.}
    \label{tab:resolution}
    \resizebox{\columnwidth}{!}{%
    \begin{tabular}{l|ccc|ccc|ccc}
        \toprule
        \multirow{2}{*}{Resolution} & \multicolumn{3}{c|}{CAMO} & \multicolumn{3}{c|}{COD10K} & \multicolumn{3}{c}{NC4K} \\
        \cline{2-10}
        & $S_\alpha$ & $F_\beta^w$ & $\mathcal{M}$ & $S_\alpha$ & $F_\beta^w$ & $\mathcal{M}$ & $S_\alpha$ & $F_\beta^w$ & $\mathcal{M}$ \\
        \midrule
        $384 \times 384$ & 0.878 & 0.847 & 0.041 & 0.873 & 0.801 & 0.021 & 0.892 & 0.857 & 0.030 \\
        $512 \times 512$ & \textbf{0.887} & \textbf{0.870} & \textbf{0.037} & 0.890 & 0.839 & 0.020 & 0.895 & 0.860 & \textbf{0.025} \\
        $1024 \times 1024$ & 0.884 & 0.860 & 0.041 & \textbf{0.908} & \textbf{0.867} & \textbf{0.016} & \textbf{0.903} & \textbf{0.877} & 0.028 \\
        \bottomrule
    \end{tabular}%
    }
\end{table}

\begin{figure}[t]
    \centering
    \includegraphics[width=\columnwidth]{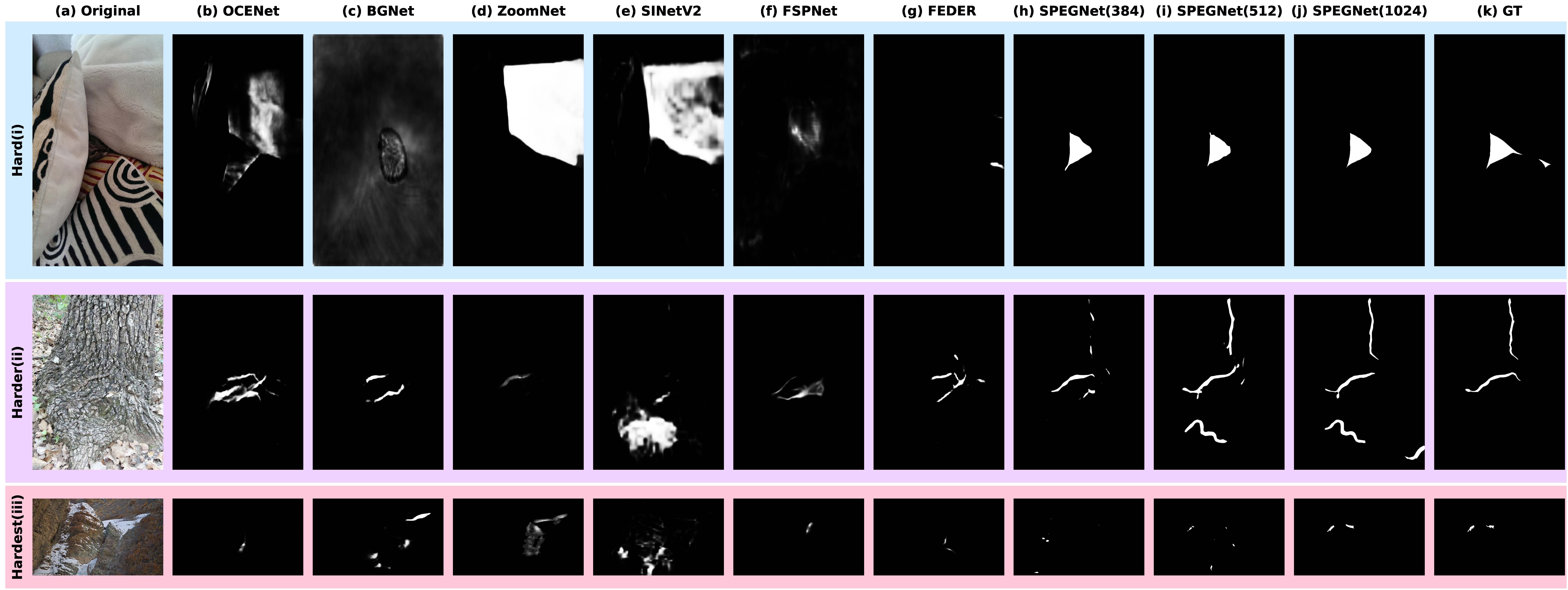}
    \caption{Resolution impact across complexity levels. Rows show: (i) contextual complexity, (ii) texture integration, and (iii) perceptual limits. Columns: (a) input, (b-g) competing methods, (h-j) SPEGNet at different resolutions, (k) ground truth.}
    \label{fig:resolution_impact}
\end{figure}

\paragraph{Visual Analysis} Fig.~\ref{fig:resolution_impact} illustrates the effects of resolution across three complexity levels, forming a perceptual spectrum ordered by increasing difficulty. Initially, Fig.~\ref{fig:resolution_impact} (row i) shows contextual complexity, where objects maintain distinct features but require complete scene understanding. Here, SPEGNet successfully segments objects across all resolutions while state-of-the-art methods fail at basic localization. Subsequently, Fig.~\ref{fig:resolution_impact} (row ii) presents texture-integrated cases where objects visually merge with similar background patterns. Critical discriminative features are progressively lost during downsampling. At 384$\times$384, SPEGNet produces partial segmentation as boundary cues degrade. However, a 512$\times$512 resolution enables sufficient detail preservation for accurate delineation, sometimes exceeding the completeness of the ground truth. Furthermore, a 1024$\times$1024 resolution provides additional fine-grained features, enhancing multi-instance detection significantly better than the ground truth. Finally, Fig.~\ref{fig:resolution_impact} (row iii) illustrates perceptually indiscernible objects where minimal visual signals distinguish foreground from background. Essential discriminative features are eliminated at lower resolutions. Only SPEGNet at 1024$\times$1024 preserves these subtle cues, successfully identifying camouflaged subjects, whereas both competing methods and lower-resolution variants fail completely to localize objects. This graduated performance across the complexity spectrum demonstrates our architecture's capacity to effectively leverage high-resolution information, addressing fundamental limitations in current approaches that sacrifice visual fidelity for computational efficiency.

\noindent The complexity spectrum maps directly to operational scenarios requiring adaptive real-time processing. Low-complexity targets need rapid area coverage while high-complexity cases demand maximum resolution. Table~\ref{tab:resolution} demonstrates SPEGNet maintains 58-63 FPS across all resolutions. This minimal variation enables seamless resolution switching during operation. Military surveillance could alternate between terrain scanning at 512$\times$512 and threat confirmation at 1024$\times$1024. Wildlife surveys could process streams continuously while adapting to target complexity. Medical procedures requiring live feedback could increase magnification for suspicious areas. The 1.25ms total scaling from lowest to highest resolution preserves operational continuity. Real-time applications can select resolution based on detection requirements rather than speed limitations. This flexibility addresses deployment scenarios where both coverage and detail matter.

\subsection{Ablation Studies}

\subsubsection{Component Analysis}

\begin{figure}[t]
    \centering
    \includegraphics[width=\columnwidth]{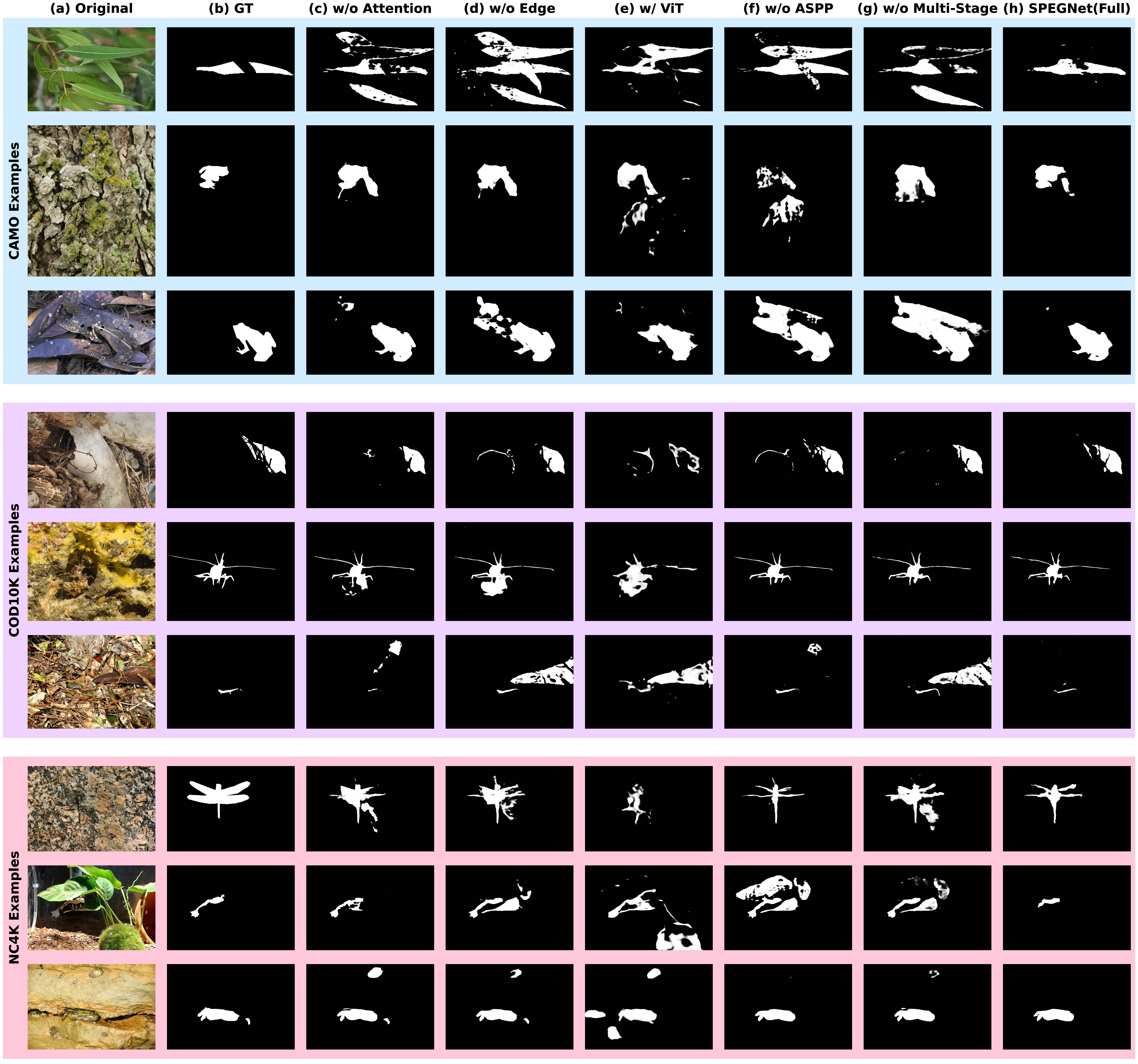}
    \caption{Visual ablation analysis on challenging examples. Columns: (a) input, (b) ground truth, (c) w/o Channel Attention, (d) w/o Edge Guidance, (e) w/ ViT, (f) w/o ASPP, (g) Single-stage Decoder, (h) SPEGNet (Full). Each variant shows specific failure modes, validating component contributions.}
    \label{fig:ablation}
\end{figure}

Table~\ref{tab:ablation} presents comprehensive ablation experiments validating each architectural component's contribution through systematic removal. All variants use identical training protocols for fair comparison. Fig.~\ref{fig:ablation} provides visual evidence through representative examples from benchmark datasets demonstrating specific failure modes when components are removed.

\begin{table}[t]
    \centering
    \caption{Ablation study on SPEGNet components.}
    \label{tab:ablation}
    \resizebox{\columnwidth}{!}{%
    \begin{tabular}{l|cc|cc|cc|cc}
        \toprule
        \multirow{2}{*}{Variant} & \multicolumn{2}{c|}{CAMO} & \multicolumn{2}{c|}{COD10K} & \multicolumn{2}{c|}{NC4K} & \multicolumn{2}{c}{Model Stats} \\
        \cline{2-9}
        & $S_\alpha$ & $\mathcal{M}$ & $S_\alpha$ & $\mathcal{M}$ & $S_\alpha$ & $\mathcal{M}$ & GMac & Params(M) \\
        \midrule
        w/ ViT & 0.849 & 0.051 & 0.831 & 0.032 & 0.856 & 0.034 & \textbf{223.11} & 307.89 \\
        w/o Channel Att. & 0.872 & 0.042 & 0.875 & 0.025 & 0.881 & 0.029 & 292.10 & 215.44 \\
        w/o e-ASPP & 0.869 & 0.044 & 0.872 & 0.026 & 0.878 & 0.030 & 301.93 & 216.06 \\
        w/o Edge Guidance & 0.865 & 0.045 & 0.868 & 0.028 & 0.875 & 0.031 & 284.24 & 215.07 \\
        Single-stage Dec. & 0.871 & 0.043 & 0.873 & 0.027 & 0.879 & 0.030 & 241.10 & \textbf{214.00} \\
        \midrule
        SPEGNet (Full) & \textbf{0.887} & \textbf{0.037} & \textbf{0.890} & \textbf{0.020} & \textbf{0.895} & \textbf{0.025} & 292.10 & 215.44 \\
        \bottomrule
    \end{tabular}%
    }
\end{table}

\paragraph{Encoder Impact} Substituting hierarchical Hiera backbone with standard ViT demonstrates substantial performance degradation (6.6\% $S_\alpha$ drop on COD10K). Despite this architectural mismatch, the ViT variant maintains reasonable effectiveness ($S_\alpha$ scores above 0.831), confirming SPEGNet's fundamental soundness independent of encoder choice. Furthermore, the ViT variant's decreased boundary precision is particularly evident in challenging scenarios that require fine-grained discrimination. Fig.~\ref{fig:ablation} rows 1 and 7 show how complex structural details in insects and dragonflies are less accurately preserved with the ViT backbone. Moreover, the ViT configuration reduces computational cost (223.11 vs. 292.10 GMac) while increasing the number of parameters (307.89M vs. 215.44M); however, it suffers from both slower inference and degraded boundary precision. This validates our choice of hierarchical transformer backbone for effective camouflage detection.

\paragraph{Channel Attention} Removing channel attention yields consistent performance degradation across datasets (1.7\% $S_\alpha$ drop on COD10K). The impact is most pronounced in complex scenarios with substantial similarity between the background objects. Fig.~\ref{fig:ablation}(c) (row 1) shows how, without channel attention, the model incorrectly identifies background leaves as insect parts, demonstrating reduced discriminative capability despite preserving main object structure. Similarly, Fig.~\ref{fig:ablation}(c) (row 2) shows additional structures near the creature's head incorrectly included in segmentation. These observations confirm that channel attention effectively suppresses similarly-textured background features while enhancing object-specific channels.

\paragraph{Multi-scale Context} Eliminating e-ASPP reduces effectiveness (2\% $S_\alpha$ drop on COD10K) while increasing computational cost (301.93 GMac). Fig.~\ref{fig:ablation}(f) (row 3) illustrates how the model without ASPP fails to capture the correct frog shape, instead detecting a larger, incorrectly bounded region due to inadequate multi-scale context integration. This confirms e-ASPP's role in capturing scale-dependent camouflage properties through parallel receptive fields of varying sizes.

\paragraph{Edge Guidance} This component shows strongest impact (2.5\% $S_\alpha$ drop on COD10K). The structural score degradation reveals how object coherence collapses when edge guidance is removed. Fig.~\ref{fig:ablation}(d) illustrates this collapse across different scenarios. Fig.~\ref{fig:ablation}(d) (row 1) shows how the model incorrectly segments multiple leaf regions as insect parts while simultaneously losing fine structural details. Additionally, Fig.~\ref{fig:ablation}(d) (row 4) shows edge-free processing struggling with occlusion handling, failing to maintain object completeness while introducing boundary noise. Fig.~\ref{fig:ablation}(d) (row 6) demonstrates massive over-segmentation consuming entire regions. These failures confirm edge guidance contributes beyond simple boundary enhancement, enabling accurate object localization through texture-boundary discrimination.

\paragraph{Progressive Refinement} Single-stage decoder achieves lowest computational cost (241.10 GMac) but significantly impacts performance (1.9\% $S_\alpha$ drop on COD10K). Fig.~\ref{fig:ablation}(g) (row 7) demonstrates how single-stage processing cannot achieve the same level of structural refinement in dragonfly wing patterns compared to our whole model. Similarly, Fig.~\ref{fig:ablation}(g) (row 2) shows that while capturing basic object shape, the single-stage variant lacks refined structural details. This validates our scale-dependent edge influence approach, where graduated refinement mirrors Bayesian updating, with coarser predictions serving as priors progressively updated by finer-scale evidence.

\subsubsection{Computational Analysis}

Table~\ref{tab:model_comparison} compares computational characteristics where SPEGNet operates at a higher 512$\times$512 resolution while FEDER and FSPNet use 384$\times$384. Despite processing 78\% more pixels, SPEGNet maintains competitive inference speed. We additionally show SPEGNet scaling from 384×384 to 1024×1024.

\paragraph{Speed at Higher Resolution} SPEGNet achieves 16.50ms inference at 512×512—our standard evaluation resolution. This surpasses all competitors in Table~\ref{tab:comparison}, while processing significantly more pixels than FEDER and FSPNet at 384$\times$384. FEDER's 81.65ms latency at lower resolution demonstrates that minimal FLOPs (36.05 GMac) fail to ensure practical speed. Sequential frequency-domain processing creates bottlenecks independent of theoretical efficiency. Conversely, SPEGNet's 16.50ms at 512$\times$512 enables real-time deployment, preserving camouflage details.

\paragraph{Fair Resolution Comparison} At equal 384$\times$384 resolution, SPEGNet requires only 15.75ms—comparable to FSPNet (15.35ms) despite 5$\times$ more FLOPs. This validates the advantages of transformer parallelization for COD. However, a 384$\times$384 resolution sacrifices critical features, as shown in Fig.~\ref{fig:resolution_impact}. SPEGNet's ability to maintain real-time speed at 512$\times$512 and 1024$\times$1024 addresses this fundamental limitation. The 1.25ms scaling from 384$\times$384 to 1024$\times$1024 demonstrates efficient multi-resolution processing.

\paragraph{Performance and Resolution Trade-off} SPEGNet achieves superior detection (0.890 $S_\alpha$ on COD10K) at 512$\times$512 while competitors constrain to 384×384. This 78\% resolution advantage directly contributes to performance gaps. Our architecture comprises 212.15M from the hierarchical backbone and only 3.29M (1.5\%) from our synergistic modules. This focused design enables high-resolution processing that FEDER's (44.13M parameters) cannot achieve despite theoretical efficiency. FSPNet's (274.24M parameters) exhibit diminishing returns, resulting in lower performance at lower resolutions. Real-world deployment benefits from SPEGNet's resolution flexibility, matching camouflage complexity.

\begin{table}[t]
    \centering
    \caption{Computational performance comparison. FEDER and FSPNet operate at 384$\times$384, while SPEGNet displays results at three different resolutions.}
    \label{tab:model_comparison}
    \resizebox{\columnwidth}{!}{%
    \begin{tabular}{l|ccc}
        \toprule
        Methods & Inference (ms) & FLOPs (GMac) & Params (M) \\
        \midrule
        FEDER-384 & 81.65 & 36.05 & 44.13 \\
        FSPNet-384 & 15.35 & 281.05 & 274.24 \\
        \midrule
        SPEGNet-384 & 15.75 & 186.19 & 215.44 \\
        SPEGNet-512 & 16.50 & 292.10 & 215.44 \\
        SPEGNet-1024 & 17.00 & 1168.40 & 215.44 \\
        \bottomrule
    \end{tabular}%
    }
\end{table}

\subsection{Application to Related Domains}

\begin{figure}[t]
    \centering
    \includegraphics[width=\columnwidth]{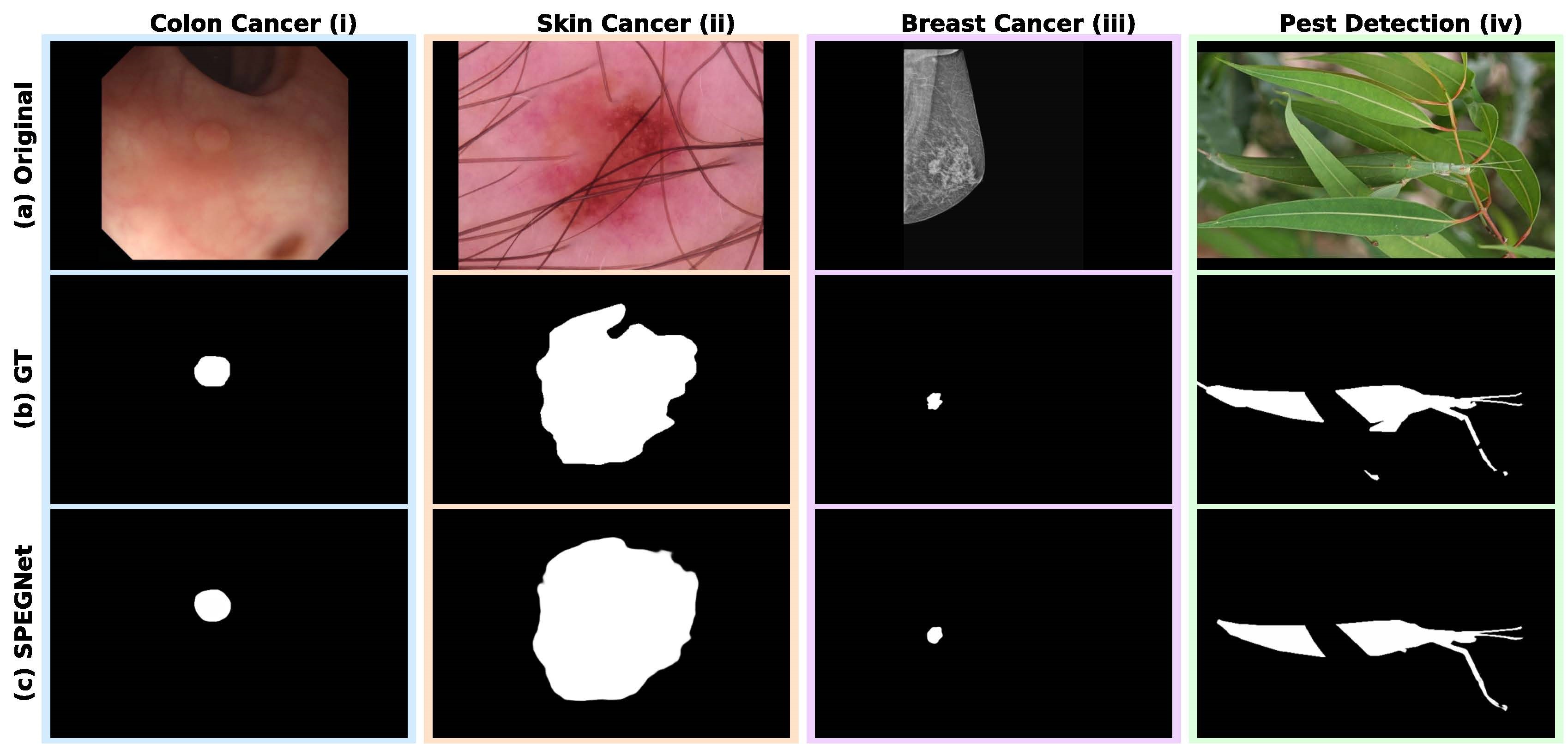}
    \caption{Application of SPEGNet to medical imaging and agricultural domains. Columns show different domains: (i) Colon Cancer - polyp blending with intestinal tissue, (ii) Skin Cancer - lesion with unclear boundaries, (iii) Breast Cancer - low-contrast mammogram lesion, (iv) Pest Detection - locust camouflaged on plant leaves. Rows show: (a) Original image, (b) Ground truth, (c) SPEGNet prediction. The model successfully identifies targets exhibiting camouflage-like properties across all domains without architectural modifications.}
    \label{fig:applications}
\end{figure}

To demonstrate SPEGNet's generalization beyond natural camouflage, we evaluate direct application to medical imaging and agricultural domains without architectural modifications. Fig.~\ref{fig:applications} illustrates representative examples where targets exhibit camouflage-like properties: polyps blending with intestinal tissue, skin lesions with unclear boundaries, breast lesions in low-contrast mammograms, and pests mimicking plant textures. Table~\ref{tab:applications} presents a quantitative evaluation across multiple datasets.

\begin{table*}[t]
    \centering
    \caption{SPEGNet performance on medical imaging and agricultural applications without domain-specific modifications. Best results in \textbf{bold}.}
    \label{tab:applications}
    \resizebox{\textwidth}{!}{%
    \begin{tabular}{l|ccccc|ccc|cc|c}
        \toprule
        & \multicolumn{5}{c|}{\textbf{Colon Polyp Detection (mDice$\uparrow$)}} & \multicolumn{3}{c|}{\textbf{Skin Lesion (mDice$\uparrow$)}} & \multicolumn{2}{c|}{\textbf{Breast Lesion}} & \textbf{Pest (mDice$\uparrow$)} \\
        \cmidrule(lr){2-6} \cmidrule(lr){7-9} \cmidrule(lr){10-11} \cmidrule(l){12-12}
        Method & Kvasir & CVC-Clinic & ColonDB & CVC300 & ETIS & ISIC & PH2 & HAM10K & BCSD (IoU$\uparrow$) & DMID (mDice$\uparrow$) & Locust-mini \\
        \midrule
        SPEGNet & 0.927 & 0.905 & \textbf{0.792} & 0.892 & 0.748 & 0.864 & 0.953 & \textbf{0.914} & 0.745 & \textbf{0.707} & \textbf{0.930} \\
        \midrule
        SOTA & \textbf{0.939} & \textbf{0.961} & 0.776 & \textbf{0.906} & \textbf{0.879} & \textbf{0.987} & \textbf{0.954} & 0.906 & \textbf{0.750} & 0.660 & 0.861 \\
        Model & \multicolumn{5}{c|}{RAPUNet~\cite{lee2024RAPUNet}} & \multicolumn{3}{c|}{MFSNet~\cite{basak2022mfsnet}} & SAM~\cite{sam2023} & Att U-Net~\cite{oktay2018attention} & Polyp-PVT~\cite{dong2023PolypPVT} \\
        \bottomrule
    \end{tabular}%
    }
\end{table*}

\paragraph{Medical Imaging Performance} SPEGNet demonstrates competitive performance on medical segmentation tasks, sharing visual similarity challenges with camouflage. For colon polyp detection (Fig.~\ref{fig:applications}(i)), the small polyp blends seamlessly with surrounding tissue, yet our SPEGNet accurately segments it. SPEGNet achieves 0.792 mDice on ColonDB, improving upon RAPUNet~\cite{lee2024RAPUNet}'s 0.776 by 2.1\%. On Kvasir and CVC-Clinic datasets, we achieve competitive performance with 0.927 and 0.905 mDice respectively. Skin lesion segmentation (Fig.~\ref{fig:applications} column ii) shows SPEGNet capturing irregular boundaries despite texture similarity, achieving 0.914 on HAM10K, surpassing MFSNet~\cite{basak2022mfsnet}. Breast lesion detection (Fig.~\ref{fig:applications}(iii)) in low-contrast mammograms yields 0.707 mDice on DMID, significantly outperforming Attention U-Net~\cite{oktay2018attention}.

\paragraph{Agricultural Transfer Results} Pest detection demonstrates exceptional cross-domain transfer. Fig.~\ref{fig:applications}(iv) shows a locust perfectly camouflaged against plant leaves—a natural camouflage scenario directly analogous to our training domain. SPEGNet precisely segments the pest, achieving 0.930 mDice on the Locust-mini dataset, substantially outperforming Polyp-PVT~\cite{dong2023PolypPVT}'s 0.861. This 8.0\% improvement validates that our edge-guided refinement and multi-scale processing effectively handle agricultural pest detection. The visual similarity between natural camouflage and pest-crop blending explains this strong performance.

\paragraph{Camouflage Principles in Other Domains} The visual examples in Fig.~\ref{fig:applications} reveal why SPEGNet transfers effectively: each domain exhibits core camouflage properties of intrinsic similarity and edge disruption. Medical lesions blend through color/texture matching, while pests evolved natural camouflage. SPEGNet's synergistic design captures these fundamental principles of visual discrimination across all domains. The consistent architecture effectively handles medical and agricultural applications. Future work will explore domain-specific fine-tuning to bridge the remaining performance gaps with specialized models. While these specialized models currently achieve marginally higher scores through domain engineering, SPEGNet provides a unified solution across all applications. This direct applicability without architectural adaptation validates our approach to addressing visual similarity challenges that transcend specific camouflage contexts.

\section{Analysis and Discussion}
\label{sec:analysis}

This section offers in-depth insights into camouflage detection challenges through a detailed analysis of the dataset. We examine specific failure modes, annotation inconsistencies, and fundamental detection limits. This analysis reveals both the effectiveness of our approach and the inherent challenges in the field.

\subsection{Dataset-Specific Challenges and Model Behavior}

\begin{figure}[t]
    \centering
    \includegraphics[width=\columnwidth]{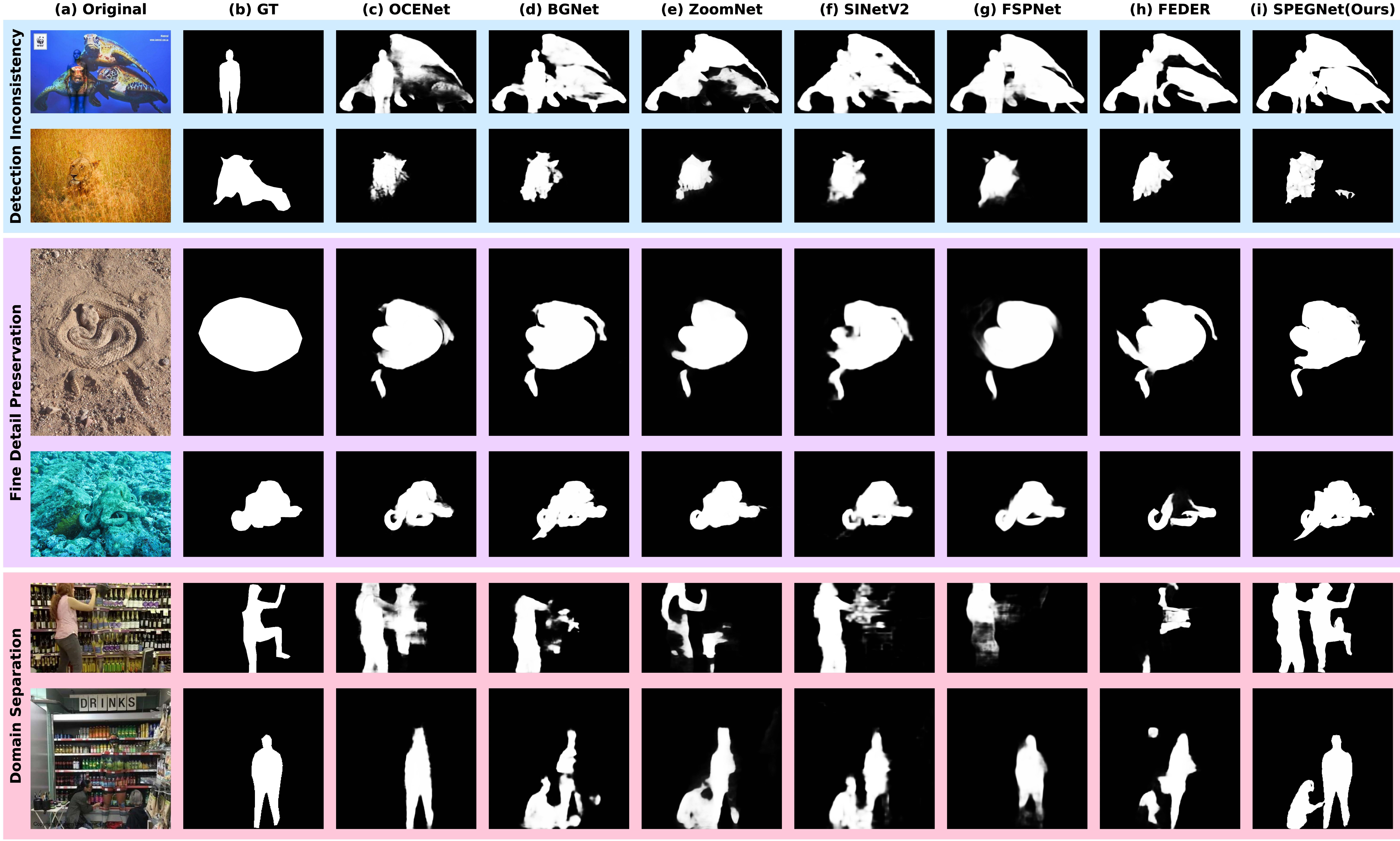}
    \caption{Analysis of CAMO dataset challenges. Columns show: (a) Input images, (b) Ground truth masks, and predictions from (c) OCENet, (d) BGNet, (e) ZoomNet, (f) SINetV2, (g) FSPNet, (h) FEDER, and (i) SPEGNet. Rows demonstrate: artificial camouflage (row 1), occlusion handling (row 2), boundary detail preservation (rows 3-4), and instance segmentation complexity (rows 5-6).}
    \label{fig:camo_analysis}
\end{figure}

Fig.~\ref{fig:camo_analysis} presents representative challenges from the CAMO dataset showcasing different detection scenarios. Fig.~\ref{fig:camo_analysis} (row 1) demonstrates artificial camouflage patterns where human subjects blend with designed backgrounds. This scene also illustrates the Salient-Camouflaged Object Disambiguation (SCOD) challenge—a fundamental problem in COD where scenes contain both prominent salient objects and subtle camouflaged objects, requiring models to detect only the camouflaged target while treating salient objects as background despite their visual prominence. The ground truth marks only the camouflaged human while ignoring visually distinctive turtle patterns. SPEGNet, like other methods without explicit saliency modeling, segments both patterns as the visual distinction requires understanding camouflage-saliency relationships beyond appearance features. Fig.~\ref{fig:camo_analysis} (row 2) illustrates occlusion handling capabilities where environmental elements obscure parts of organisms. Our progressive refinement approach maintains object completeness better than competing methods. Fig.~\ref{fig:camo_analysis} rows 3-4 examine boundary detail preservation in elongated organisms where fine anatomical structures challenge detection accuracy. SPEGNet's edge-guided decoder captures these details more precisely than alternatives. Fig.~\ref{fig:camo_analysis} rows 5-6 address instance segmentation complexity in multi-object scenarios. These cases reveal systematic annotation limitations where ground truth marks fewer instances than are actually present, creating evaluation ambiguities.

\subsection{Extreme Camouflage and Context Dependencies}

\begin{figure}[t]
    \centering
    \includegraphics[width=\columnwidth]{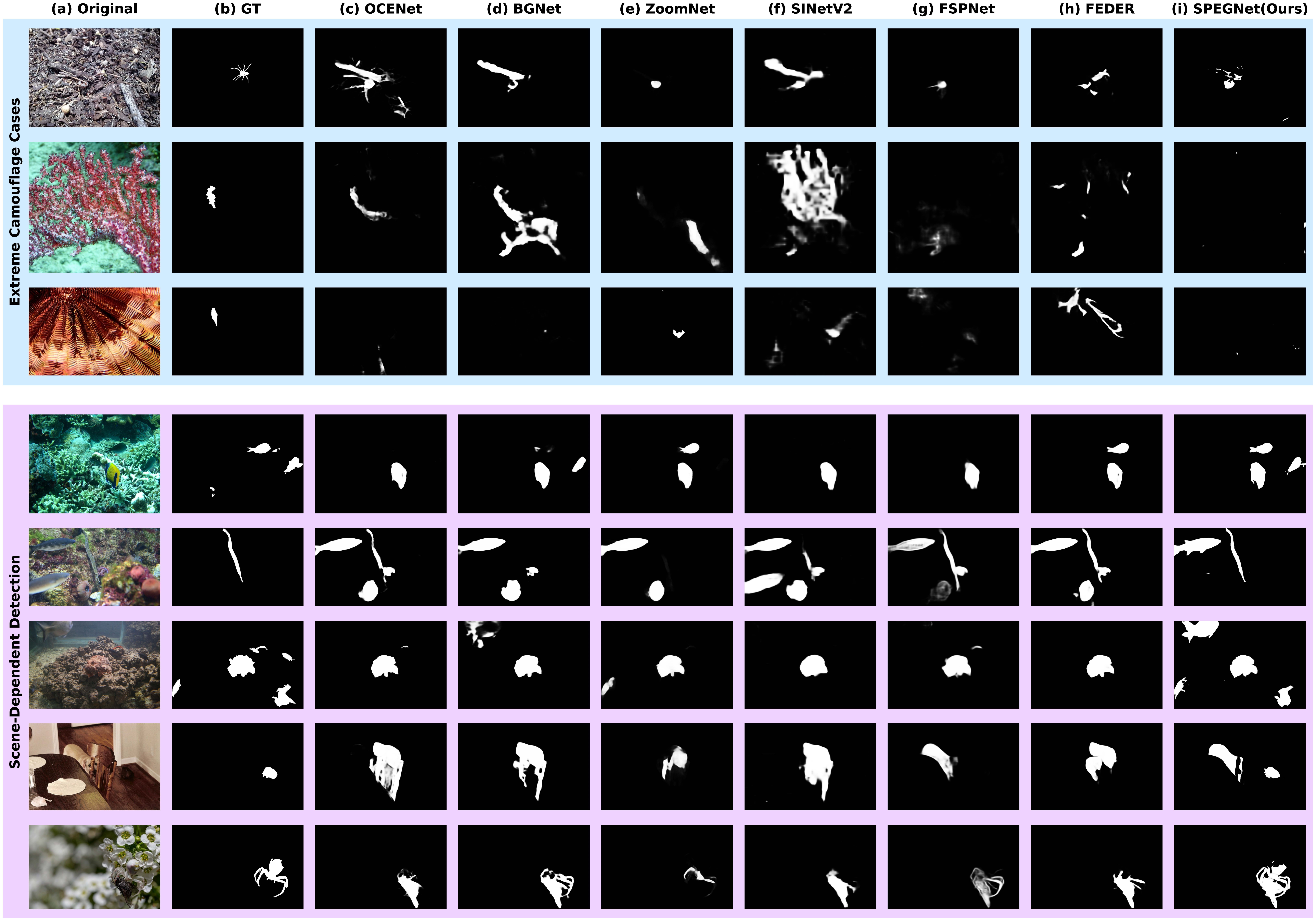}
    \caption{Analysis on COD10K's challenging cases. Columns show: (a) Input images, (b) Ground truth masks, and predictions from (c) OCENet, (d) BGNet, (e) ZoomNet, (f) SINetV2, (g) FSPNet, (h) FEDER, and (i) SPEGNet. The top three rows show extreme camouflage cases where objects are barely visible. The bottom five rows illustrate scene-dependent camouflage, where similar objects are labeled differently depending on the scene context.}
    \label{fig:cod10k_analysis}
\end{figure}

\paragraph{Perceptual Detection Limits} Fig.~\ref{fig:cod10k_analysis} reveals fundamental boundaries in camouflage detection through COD10K examples. Fig.~\ref{fig:cod10k_analysis} top three rows present extreme camouflage cases where objects are barely visible even to human observers. These organisms achieve near-perfect background integration through evolutionary adaptations in texture, pattern, and color. At standard processing resolutions, all current architectures, including SPEGNet, fail to detect these objects as discriminative features become spatially indistinguishable. This observation directly connects to our resolution analysis, where increasing the resolution to 1024$\times$1024 enables the detection of previously undetectable instances. The consistent failure across all methods suggests this limitation transcends specific architectural designs, representing a fundamental resolution-dependent bottleneck.

\paragraph{Scene-Dependent Annotation Logic} The bottom five rows of Fig.~\ref{fig:cod10k_analysis} demonstrate scene-dependent camouflage where similar objects receive different annotations based on scene context rather than intrinsic visual properties. Taxonomically identical organisms are marked as camouflaged or background, depending entirely on their surrounding environment. This context dependency presents challenges beyond visual appearance—models must understand environmental relationships to match human annotation logic. SPEGNet, like other current methods that rely on pre-trained visual encoders, tends to detect all visually similar objects regardless of their contextual camouflage status. This consistent behavior stems from an architectural focus on appearance-based features rather than scene-level reasoning.

\subsection{Cross-Dataset Generalization Patterns}

\begin{figure}[t]
    \centering
    \includegraphics[width=\columnwidth]{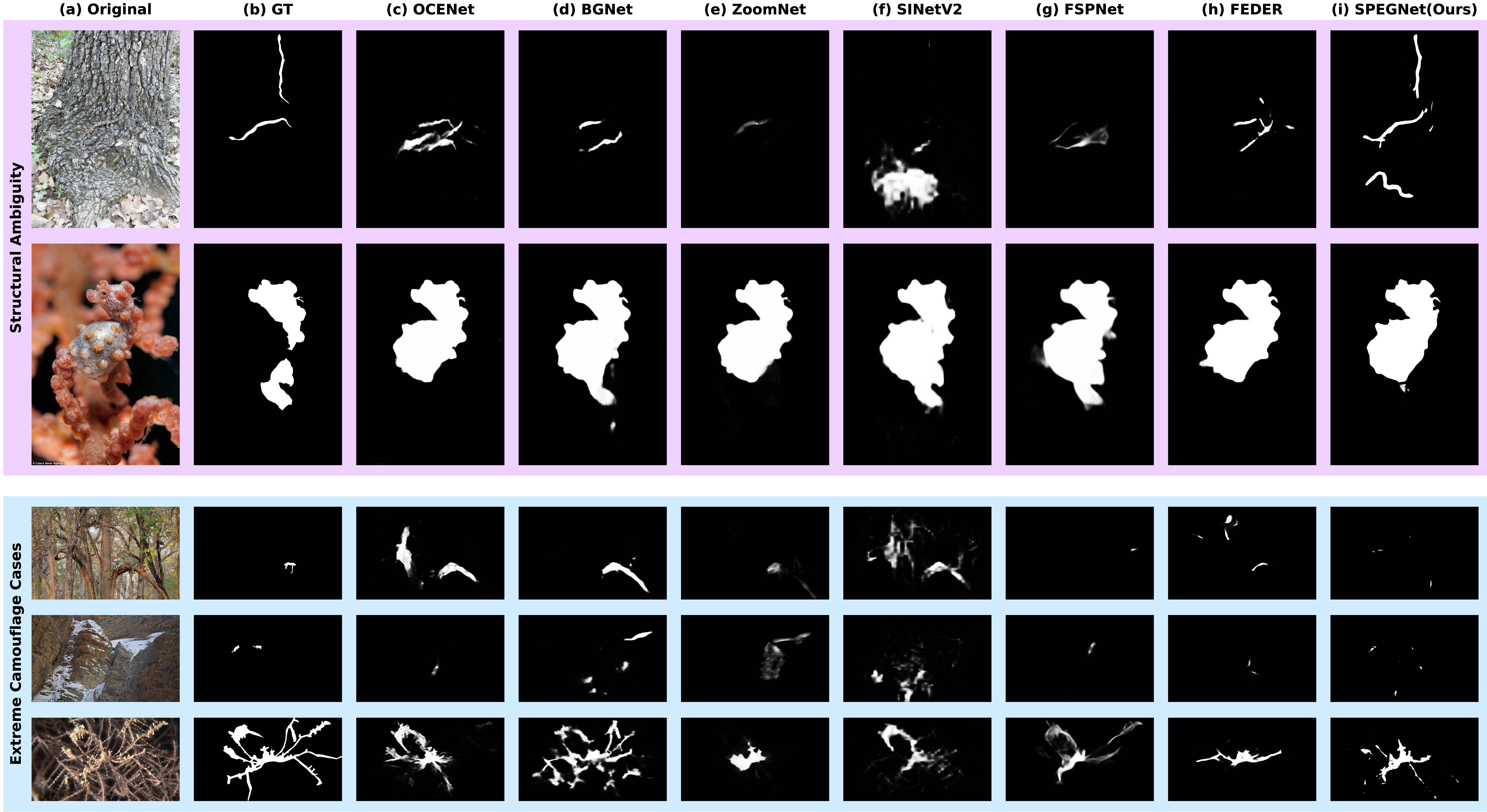}
    \caption{Analysis of NC4K test cases. Columns show: (a) Input images, (b) Ground truth masks, and predictions from (c) OCENet, (d) BGNet, (e) ZoomNet, (f) SINetV2, (g) FSPNet, (h) FEDER, and (i) SPEGNet. The top two rows show structural ambiguity cases where models' GT agreement is complex. The bottom three rows present extreme camouflage cases with highly challenging background similarity.}
    \label{fig:nc4k_analysis}
\end{figure}

\paragraph{Structural Ambiguity Cases} Fig.~\ref{fig:nc4k_analysis} examines SPEGNet's performance on the unseen NC4K benchmark. Fig.~\ref{fig:nc4k_analysis} top two rows present structural ambiguity cases where model-ground truth agreement becomes complex. These scenarios involve organisms with discontinuous parts or unclear boundary definitions that could be considered integral to the target. SPEGNet demonstrates the capability to capture morphological details that competing methods miss, though this sometimes exceeds ground truth annotations. Fig.~\ref{fig:nc4k_analysis} bottom three rows show extreme camouflage cases with highly challenging background similarity requiring maximum discriminative capability.

\paragraph{Zero-Shot Transfer Validation} Despite never encountering NC4K during training, SPEGNet maintains state-of-the-art performance (0.895 $S_\alpha$) demonstrating robust generalization capabilities. The consistent performance patterns across this unseen benchmark validate that our synergistic design learns fundamental camouflage principles rather than dataset-specific patterns. Our architecture's effective balance of feature integration, context modeling, and edge-guided refinement enables robust transfer to novel camouflage instances while maintaining shared limitations in extremely challenging cases.

\subsection{Component Contribution Analysis}

\begin{figure}[t]
    \centering
    \includegraphics[width=\columnwidth]{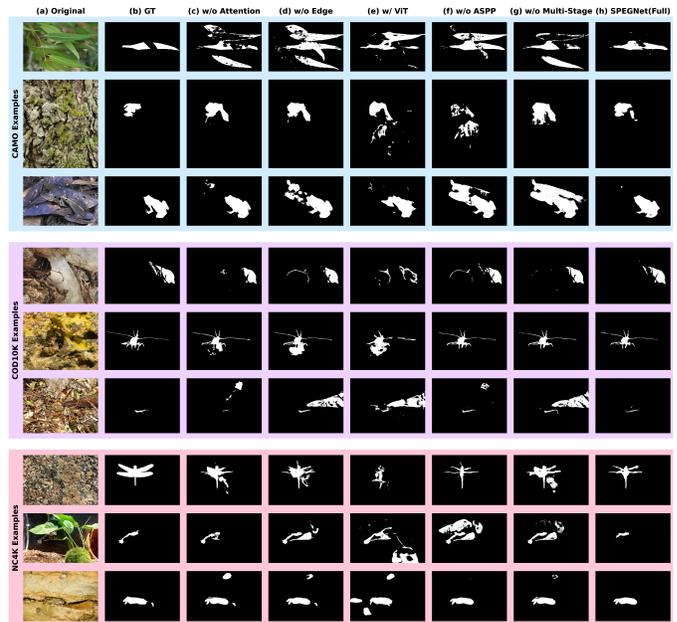}
    \caption{Visual comparison of ablation variants on challenging camouflage examples from CAMO, COD10K, and NC4K datasets. Columns show: (a) Original image, (b) Ground truth, (c) w/o Channel Attention, (d) w/o Edge Guidance, (e) w/ ViT, (f) w/o ASPP, (g) Single-stage Decoder, (h) SPEGNet (Full). Each variant shows specific failure modes, validating component contributions.}
    \label{fig:ablation_visual}
\end{figure}

Fig.~\ref{fig:ablation_visual} provides a comprehensive visual analysis across nine challenging examples from all three datasets. Each ablation variant exhibits distinct failure patterns, validating the necessity of its component. Without channel attention, Fig.~\ref{fig:ablation_visual}(c), models incorrectly segment background regions sharing similar textures with targets. Removing edge guidance, Fig.~\ref{fig:ablation_visual}(d), causes severe over-segmentation and loss of fine structural details. The ViT variant in Fig.~\ref{fig:ablation_visual}(e) maintains reasonable detection capability but suffers from reduced boundary precision in complex scenarios. Eliminating ASPP in Fig.~\ref{fig:ablation_visual}(f) leads to incorrect shape detection and inadequate multi-scale context integration. Single-stage decoder shown in Fig.~\ref{fig:ablation_visual}(g) captures basic object shapes but lacks the refined structural details achieved through progressive refinement.

\subsection{Limitations and Future Directions}

\paragraph{Resolution-Dependent Detection Boundaries} Analysis reveals fundamental limits when discriminative features become spatially compressed beyond detection thresholds. Extreme camouflage cases, representing 10-15\% of challenging instances, require resolutions exceeding current computational feasibility. These perceptual threshold cases suggest architectural improvements alone cannot address all detection challenges—alternative sensing modalities or temporal information may be necessary.

\paragraph{SCOD Challenge Requirements} The Salient-Camouflaged Object Disambiguation challenge remains unaddressed by current COD architectures, including SPEGNet. Standard COD benchmarks lack salient object annotations, which prevents a fair evaluation of SCOD-aware methods. Existing approaches either use external priors through multi-task frameworks or require saliency supervision, introducing computational overhead and external dependencies. Future architectures must develop intrinsic SCOD mechanisms without external dependencies—potentially through contrastive learning between camouflaged and salient objects or attention mechanisms that explicitly model visual prominence versus background integration.

\paragraph{Annotation Quality Challenges} Systematic inconsistencies across datasets create evaluation artifacts affecting performance assessment. Instance segmentation complexity in multi-object scenes, structural ambiguity in boundary definitions, and context-dependent labeling introduce measurement noise. These annotation limitations significantly bias evaluation metrics. As demonstrated in our experiments, CAMO's single-instance annotations penalize methods for correctly detecting multiple valid objects (Fig.~\ref{fig:visual_comparison} row iii). Such systematic biases artificially lower scores for more capable models. Future COD benchmarks urgently require exhaustive multi-instance labeling, clear protocols for ambiguous boundaries, and standardized annotation guidelines to enable fair model comparison.
\section{Conclusion}
\label{sec:conclusion}

Camouflaged object detection faces fundamental challenges due to intrinsic similarity and edge disruption. Current methods fragment detection by accumulating complex components without proportional gains. This accumulation forces reduced resolution processing, eliminating fine details essential for camouflage detection. We presented SPEGNet, addressing fragmentation through synergistic design principles. Our architecture integrates three complementary modules that work in concert rather than in parallel, allowing for accumulation. CFI combines channel recalibration with spatial enhancement to extract discriminative features. EFE derives boundaries directly from context-rich representations, maintaining semantic-spatial alignment. PED implements scale-adaptive edge modulation with peak influence at intermediate resolutions. This synergistic approach achieves state-of-the-art performance: 0.890 $S_\alpha$ on COD10K, 0.895 on NC4K, and 0.887 on CAMO with 16.5ms inference speed.

Beyond camouflage detection, SPEGNet demonstrates broad applicability to medical imaging and agricultural domains without architectural modifications. The synergistic design captures fundamental visual discrimination principles transferring across applications where targets exhibit similarity-based concealment. Our analysis reveals systematic challenges, including SCOD scenarios and resolution-dependent detection boundaries, requiring future architectural innovations. The synergistic paradigm establishes foundations for next-generation detection architectures moving beyond component accumulation toward principled integration.

\section*{Acknowledgment}
The authors would like to thank King Fahd University of Petroleum and Minerals (KFUPM) for providing the institutional support and resources that made this research possible. Special recognition goes to the Joint Research Center for Artificial Intelligence (JRCAI) at KFUPM for providing the computational resources, infrastructure, and collaborative environment essential for conducting this work. The authors also acknowledge the valuable discussions and feedback from colleagues at KFUPM that helped shape this research.

\bibliographystyle{IEEEtran}
\bibliography{references}


\end{document}